%% file: main_updated_new_solutions.tex
\documentclass{article} 

\usepackage{natbib, times}

\usepackage{geometry}
\input{math_commands.tex}

\usepackage{amsthm}
\usepackage{mathtools}
\newtheorem{question}{Question}
\newtheorem{conjecture}{Conjecture}

\usepackage{comment}

\newcommand{\ash}[1]{}
\newcommand{\alan}[1]{}
\newcommand{\tali}[1]{}
\newcommand{\mich}[1]{}

\usepackage{tikz}
\usetikzlibrary{arrows.meta}
  \tikzset{
    flow/.style={
      -{Stealth},
      thick
    }
  }

\usepackage{subcaption}
\usepackage{wrapfig}

\usepackage{arydshln}

\usepackage{needspace}
\usepackage{etoolbox}

\makeatletter
\newcommand{\WFskip}{\par
  \@tempcnta\c@WF@wrappedlines \advance\@tempcnta\m@ne
  \ifnum\@tempcnta>\@ne
    \vspace*{\dimexpr\@tempcnta\baselineskip-\intextsep\relax}\fi
  \WFclear}
\patchcmd{\WF@putfigmaybe}{\ifWF@float \dimen@.5\baselineskip}
  {\ifWF@float \dimen@2\baselineskip}{}
  {\PackageWarning{main}{Could not patch wrapfig}}
\patchcmd{\WF@putfigmaybe}{\global\WF@floatfalse\pagebreak}
  {\ifWF@float\ifx\@deferlist\@empty\global\WF@floatfalse\pagebreak\fi
   \else\global\WF@floatfalse\pagebreak\fi}{}
  {\PackageWarning{main}{Could not patch wrapfig}}
\makeatother

\usepackage{hyperref}
\usepackage[nameinlink,noabbrev]{cleveref}

\crefname{appendix}{appendix}{appendices}
\Crefname{appendix}{Appendix}{Appendices}
\crefname{conjecture}{conjecture}{conjectures}
\Crefname{conjecture}{Conjecture}{Conjectures}
\crefname{question}{question}{questions}
\Crefname{question}{Question}{Questions}

\crefformat{section}{\S#2#1#3}
\Crefformat{section}{\S#2#1#3}
\crefformat{subsection}{\S#2#1#3}
\Crefformat{subsection}{\S#2#1#3}

\makeatletter
\newif\if@skiplinkon
\ifdefined\pdfrunninglinkoff
\ifdefined\pdfannot@link@off@@
  \protected\def\pdfannot@link@off@@{\pdfrunninglinkoff}
  \protected\def\pdfannot@link@on@@{%
    \if@skiplinkon
      \global\@skiplinkonfalse
    \else
      \pdfrunninglinkon
      \global\@skiplinkontrue
    \fi}
\else
  \PackageWarningNoLine{main}{LaTeX kernel older than 2025-06-01:
    links broken across pages will still cover the header and footer}
\fi
\fi
\makeatother

\usepackage{url}

\newcommand{\pkg}[1]{\textnormal{\texttt{#1}}}

\title{
Ramanujan Challenge Results: AI-Assisted Evaluation Pipeline for Multi-Format Proofs
}

\author{Alan Sorani\textsuperscript{1}, Ashvni Narayanan\textsuperscript{1}, Michael Shalyt\textsuperscript{1}, Tali Monderer\textsuperscript{1},\\ Rotem Kalisch\textsuperscript{1}, Hila Barkan\textsuperscript{1}, Elyasheev Leibtag\textsuperscript{2}, Shachar Weinbaum\textsuperscript{1},\\ Carsten Schneider\textsuperscript{3}, John Campbell\textsuperscript{4}, Ido Kaminer\textsuperscript{1}}

\date{}

\begin{document}

\maketitle

\vspace{-30pt}

\begin{center}
{\footnotesize
\textsuperscript{1}Ramanujan Machine Group, Technion, Haifa, Israel. \\
\textsuperscript{2}Department of Mathematics, Computer Science and Statistics, Ghent University, Ghent, Belgium. \\
\textsuperscript{3}Research Institute for Symbolic Computation, Johannes Kepler Universit\"at Linz, Linz, Austria. \\
\textsuperscript{4}Department of Mathematics and Statistics, Toronto Metropolitan University, Toronto, Canada.
\\
}
\end{center}

\begin{abstract}
As AI's mathematical capabilities rapidly improve, meaningful benchmarks are shifting 
from 
university-level problems
and math olympiads
to 
research-level mathematics.
The Ramanujan Challenge is a benchmark comprised of research-level problems in number theory, including unpublished results and open conjectures. 
It presented concrete formulas for mathematical constants whose correctness can be checked numerically, while their proofs may require non-obvious mathematics. 
Here, we analyze the Challenge submissions, measuring the relative success of different AI models, Computer Algebra Systems (CAS), and formal tools. 
We find correct submissions that are conceptually distinct from the official solutions, and a proof of one of the published open conjectures. 
Evaluating the submissions exposed a second challenge: judging complex proofs reliably at scale.
We introduce an AI-assisted evaluation pipeline for natural language proofs, CAS code, and Lean formalizations that assesses semantic and logical correctness. 
Compared with the conventional one-shot LLM judgment, our pipeline reduces false-positive verdicts on natural language proofs from 23\% to 6\% and improves judgment accuracy on supporting code from 46\% to 58\%.
This approach provides a methodology for evaluating research-level mathematics at scale, with potential applications across mathematics and AI for science.
\end{abstract}

\section{Introduction}
\label{section:intro}

Benchmarks measuring the mathematical capabilities of AI systems are getting saturated at an accelerating pace.
School-level problem sets such as GSM8K and MATH \citep{GSM8K, MATH2021} were followed by university-level benchmarks, such as U-Math and ASyMOB \citep{U-Math, ASyMOB}, and math olympiad questions \citep{MiniF2F, OlympiadBench, omnimath}.
The frontier of LLM evaluation is now shifting towards research-level mathematics \citep{glazer2025frontiermathbenchmarkevaluatingadvanced, Humanity-Last_Exam, AbouzaidEtAl2026, peyronnet2026lemmabenchliveresearchlevelbenchmark}.
Meaningful assessment at this level requires a rare combination: problems at the research frontier whose solutions are unpublished to avoid training-data contamination - but are known privately.


In this work, we present the results of the Ramanujan Challenge for AI \citep{shalyt2026ramanujanchallengeai}, a collection of ten research-level numerically verifiable number theory problems.
Of these, 8 were proved by the authors but unpublished, and 2 are open conjectures. The collection was released as a benchmark for AI-assisted proving.
Our analysis shows that the submissions which were most successful at producing correct natural language proofs used models from the Claude and GPT families, while such proofs from submissions combining Lean with CAS outperformed those that relied on any single tool, suggesting benefits from diversified supporting evidence.

Notably, some of the complete proofs were submitted by amateur mathematicians, and one of the open conjectures had multiple correct proofs submitted, showing the potential of advanced AI to lower the barrier of entry to parts of mathematical research.
Submitters to the challenge, who obtained correct solutions include a person with no higher mathematical education and whose solution to one of the challenge problems was found to be simpler than the original solution from the authors. They also include young independent researchers with no mathematical training, one of whom solved a famous open mathematical problem prior to their participation in the challenge, and a high-school student who achieved significant results towards proving one of the challenge conjectures. The use of AI thus allows people accross broader backgrounds to access and advance research-level mathematics, democratizing mathematics and opening the door to cross-field research and contributions.


Obtaining the results required addressing a second challenge: evaluating large quantities of complex mathematical proof candidates.
The recent boom in AI-assisted mathematics \citep{alon2026remarksdisproofunitdistance, feng2026autonomousmathematicsresearch, ju2026automatedconjectureresolutionformal}, while producing some incredible results \citep{openai2026navierstokes,alpoge2026jacobian}, 
created an avalanche of mathematical claims -- making this problem widespread beyond benchmarks.
%
The Ramanujan Challenge illustrates this scale: participants sent a total of 228 submissions, containing a total of 4802 pages of natural language mathematical proofs and 1,079,380 lines of code across Lean, Python, Mathematica and Sage.
Recent mathematical benchmarks approach automated validation in different ways: fully relying on a large language model (LLM) as a judge \citep{ZhangEtAl2025}, using an LLM to check equivalence against a known ground truth \citep{peyronnet2026lemmabenchliveresearchlevelbenchmark}, or using deterministic code to verify LLM-generated answers \citep{ASyMOB}.
Other benchmarks, such as IMO-GradingBench and $\mu$-Math \citep{luong2025robustmathematicalreasoning, U-Math}, evaluate LLMs as judges of mathematical proofs and solutions, showing the difficulty of achieving high agreement with expert assessments.

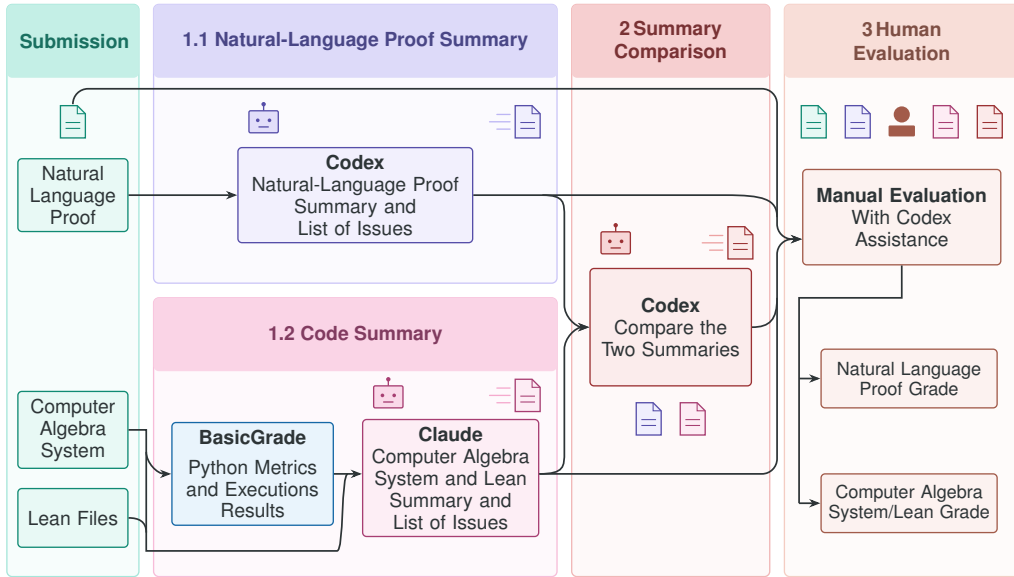
\begin{figure}[ht]
      \centering
      \def\pipelinevariant{1}
      \input{figures/manual_evaluation_color_diagram.tex}
      \caption{Overview of the evaluation pipeline.}
      \label{figure:evaluation-pipeline}
\end{figure}

In this paper, we present a novel AI-assisted pipeline, depicted in \Cref{figure:evaluation-pipeline}, for the purpose of large scale validation of
complex multi-format proofs, which include natural language, CAS code, and Lean formalizations.
We assess the Ramanujan Challenge submissions, comparing verdicts across three evaluation methods: our pipeline with and without a human expert in the loop, and a baseline LLM-as-a-judge.
Using the evaluation pipeline reduces false positive grading by a factor of 4 for natural language proofs and 11.5 for CAS/Lean code, and increases agreement with human evaluation of the code from 46\% to 58\%, relative to a baseline of the same LLM without the pipeline.

Broadly, as AI systems produce mathematical and scientific results faster than experts can verify them, validation is becoming a central bottleneck \citep{tao2026mathematicsageai,kontorovich2025shapemathcome,codreanu2026aiscienceearlyinsights}.

Our approach suggests combining natural language reasoning with complementary computational certificates, helping scale validation of AI-generated claims across mathematics and science.

\section{Related Work and Novelty}
\label{section:related-work-and-novelty}

\textbf{Related work.} Checking how well LLMs can reason is a relatively new area of interest. Last evaluated in December 2025, the Frontier Math benchmark \citep{glazer2025frontiermathbenchmarkevaluatingadvanced} considered reasoning capabilities of LLMs across multiple domains of research level mathematics. They concluded that the best models of the time were insufficient for advanced research, ``revealing a vast gap between AI capabilities and the prowess of the mathematical community''. The First Proof \citep{AbouzaidEtAl2026} benchmark focuses on unpublished research problems whose answers are known to experts. These works attest to the high cost of human feedback, and the importance of monitoring the progress of AI as a research assistant. One of the aims of this manuscript is to answer this question for current models.

The other goal is to develop an AI-assisted evaluation pipeline for given solutions. The authors of First Proof stress on the difficulty of manual evaluation when there is a large volume of submissions for complex problems. There are fewer works on the subject of using AI as graders. Using LLMs as teaching assistants for undergraduate discrete math examinations was studied in \citep{sapkota2026llmsteachingassistantsmathematics}. The paper suggests that AI can apply harsh point deductions under stricter policy conditions. LLM as a grader has also been studied in \citep{xie2024gradelikehumanrethinking}, where it is used to evaluate responses to six questions in an examination on an Operating Systems Course. Google's DeepMind constructed IMO-GradingBench in \citep{luong2025robustmathematicalreasoning}, showing that autograders coupled with reasoning from Gemini seem to correlate well with human evaluation on IMO problems.
\citep{yu2026evaluating} focuses on evaluation of handwritten college mathematics using AI.

\textbf{Novelty.}
To the best knowledge of the authors, we are the first to construct an AI-assisted evaluation pipeline as well as test it in the context of large scale evaluations of research-level computational mathematics questions. Our evaluation pipeline, when used by an LLM agent, reduces false positive grading by a factor of 4 for natural language proofs and 11.5 for CAS/Lean code, and increases agreement with human evaluation of the code from 45\% to 58\%. This demonstrates an improvement in the direction of automated AI verification of natural language complex mathematical proofs.

We further contribute to the analysis of CAS as means of proof or verification for such mathematical questions, an unexplored area of research. We analyze in particular the combined use of CAS and Lean and see that it is accompanied by correct natural language proofs more so than the use of either one in isolation, opening a door to further research on the interconnection of CAS and formal theorem assistants in AI mathematics.

\section{The Trust Spectrum}
\label{section:the-trust-spectrum} 
Our proposed paradigm for measuring the validity of  solutions to mathematical questions is the ``trust spectrum''. While theoretically most mathematical proofs can be completely formalized, in practice, trust in a mathematical proof lies on a spectrum: soft verification - such as numerical validation or probabilistic checking; intermediate verification - symbolic validation by a computer algebra system (CAS); and hard verification - machine-checked theorem proving like with Lean.

CAS rely on symbolic representations and computational rules rather than axiomatic representations of mathematical objects. They can include data produced by otherwise verifiable mathematical statements, which can be used in subsequent reasoning and proofs in the common sense of the word \citep{barendregt2005computermathematics}. Provability, however, is not always formally decidable via CAS.


\begin{wrapfigure}{r}{0.44\textwidth}
    \includegraphics[width=\linewidth]{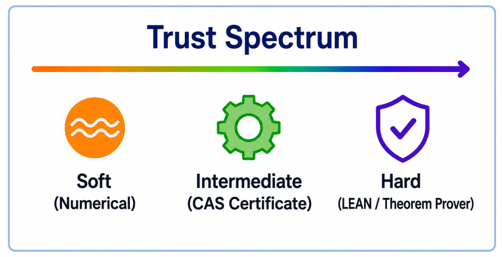}
    \vspace{-6pt}
    \caption{The trust spectrum.}
    \label{figure:trust-spectrum}
    \vspace{-8pt}
\end{wrapfigure}

ITPs such as Lean enforce strict rules for line-by-line verification of proofs. They can produce dependency graphs of the proof on other theorems, going right down to the used definitions and axioms. For this reason, we place formal verification the highest on our trust spectrum. Moreover, for a human to verify a formal proof, one needs to inspect only the definitions, used axioms, and problem statement, significantly reducing verification time.

In this paper, one of our aims is to measure how well AI systems can evaluate a given solution, using the trust spectrum as a metric. This view of trust creation in AI results can also generally help decide when to invest substantial time exploring AI outputs, and which parts of a new claim to focus on.


\section{The Challenge}
\label{section:the-challenge}
The Ramanujan Challenge focuses on explicit formulae involving important mathematical constants such as $\pi$, $e$, $\zeta(s)$ etc. The form they take are of recurrences, continued fractions, series, and integrals. Most of these formulae were discovered by exploring \textit{trajectories} arising in Conservative Matrix Fields, using the \pkg{ramanujantools} package
\citep{ramanujantools, elimelech2023PNAS,
weinbaum2025conservativematrixfieldscontinuous, raz2025PiUnifier}. 

The Ramanujan Challenge, published on arXiv \citep{shalyt2026ramanujanchallengeai} and on the Ramanujan Machine website \citep{ramanujan_challenge_site}, ran from July to August 2026. It comprised of 8 questions and 2 conjectures, listed in \Cref{appendix:challenge-questions} and \Cref{appendix:challenge-conjectures} respectively.
The solutions to these questions were (to the knowledge of the authors) previously unpublished.

The challenge asked for solutions comprising of the following data, in descending order of importance: formal proof, CAS-based derivations and natural language proofs. The reason for this order of importance is discussed in \Cref{section:the-trust-spectrum}.

At the time of posting the Challenge, the prevalent LLMs could not solve any of the problems (with one-shot prompts). The challenge received submissions from people within and outside the mathematics community, many of which did succeed in solving some of the problems.

\section{Methodology}
\label{section:methodology}
\ash{Expected: approx 1.5 pages}
In this section, we discuss the construction of our AI-assisted pipeline, described in \Cref{figure:evaluation-pipeline}. The final goal of this pipeline is to reduce human time and effort invested into largescale grading of proofs that require multiple formats.


\subsection{Evaluation Pipeline}
\label{subsection:pipeline}

For a submission, the pipeline provides a human reviewer with AI-generated summaries for each file in the submission, along with an AI-generated review connecting all of them.

The major components of the pipeline can be broken down into:
\newline
\textbf{Step 1: Basic Grade Files.}
Every Python script is executed and given a ``basic grade'' using our custom script. This is passed as input in the next step. These files comment on the successful compilation of the given code, as well as summarize the script's structure (comment density, imports, cyclomatic complexity and a keyword-based symbolic/numeric classification) and its observed behavior under a fixed runtime environment (exit status, execution time and standard output).

\textbf{Step 2: AI Summary.}
For every submission, two AI summary files are generated, one for the natural language proof and one for the code.

\textit{Natural language summary.} A GPT 5.6-Sol agent with high reasoning effort reads each PDF file in the submission folder and produces a summary file. The file contains a summary of the proof, a list of issues, and (code) files mentioned in the proof.

\textit{CAS Summary.} A Claude Opus 5 agent with high reasoning effort reads the CAS/Lean files in each submission folder as well as the basic grade files when present. It produces a file containing a summary of the code, a list of issues, and a list of files that pertain to the verification of the proof.  

Note that only relevant files are presented to the agents, not the entire submission folder. This is done to avoid contamination of ideas between the natural language and CAS/Lean segments of the proof. This is done to have a more thorough check of semantic accuracy.

\textbf{Step 3: Summary Comparison.} A GPT 5.6-Sol agent with medium reasoning effort compares the summary files produced in the previous step. It generates a new file listing the similarities between the natural language proof and the code, a list of conflicts, and a list of common issues derived from the two summary files.

\textbf{Step 4: Human Evaluation.} A human can now evaluate the submission with the aid of the summary file(s) generated above.

\begin{table}[ht]
    \centering
    \caption{Agreement of the LLM agent with human verdicts without using our pipeline (via one-shot prompting) and with the pipeline.}
    \label{table:agreement-and-false-positives}

    \begin{subtable}{\linewidth}
        \centering
        \caption{Agreement between human evaluations and those from Claude Opus 5 across submissions to the challenge questions, for natural language proofs and for CAS/Lean code, with one-shot prompting or with the evaluation pipeline.}
        \label{table:agreement}
        \small
        \begin{tabular}{l:cc:cc}
            \multicolumn{1}{c}{} & \multicolumn{2}{c}{\textbf{Natural language proof}} & \multicolumn{2}{c}{\textbf{CAS / Lean}} \\
            \\[-1.0ex]
            Verdict comparison & \shortstack{Without\\pipeline} & \shortstack{With\\pipeline} & \shortstack{Without\\pipeline} & \shortstack{With\\pipeline} \\
            \hline
            Agreement & 65.7\% & 66.4\% & 45.8\% & 58.3\% \\
            Claude stricter than human & 2.9\% & 15.3\% & 4.2\% & 29.2\% \\
            Human stricter than Claude & 31.4\% & 18.2\% & 50.0\% & 12.5\% \\
        \end{tabular}
    \end{subtable}

    \vspace{1.5em}

    \begin{subtable}{\linewidth}
        \centering
        \caption{Submissions the human evaluator graded Partial or Major Issues that Claude's verdict nonetheless called Correct, for natural language proofs and for CAS/Lean code, as a share of all graded submissions (137 proofs; 120 CAS/Lean, excluding submissions with no CAS artifact).}
        \label{table:false-positives}
        \small
        \begin{tabular}{l:ccc:ccc}
            \multicolumn{1}{c}{} & \multicolumn{3}{c}{\textbf{Natural language proof}} & \multicolumn{3}{c}{\textbf{CAS / Lean}} \\
            \\[-1.0ex]
            Human verdict & \shortstack{Number of\\submissions} & \shortstack{Without\\pipeline} & \shortstack{With\\pipeline} & \shortstack{Number of\\submissions} & \shortstack{Without\\pipeline} & \shortstack{With\\pipeline} \\
            \hline
            Partial & 30 & 23 & 7 & 65 & 39 & 4 \\
            Major Issues & 53 & 9 & 1 & 30 & 7 & 0 \\
            \hline
            Total & 83 (60.6\%) & 32 (23.4\%) & 8 (5.8\%) & 95 (79.2\%) & 46 (38.3\%) & 4 (3.3\%) \\
        \end{tabular}
    \end{subtable}
\end{table}

\subsection{Evaluation Criteria}
\label{subsection:evaluation-criteria}

We detail the evaluation criteria by format. Each submission is assigned a rank for its correctness and completeness. Correct and complete proofs or verification codes have been assessed as Correct. Partial ones that demonstrate a meaningful step towards a proof, those with a derivation gap that is not central to the entirety of the proof, or code verification for only part of the proof, have been all classified as Partial. Others with major issues or extra assumptions for the formal proofs have been classified as having Major Issues.

\textbf{Natural language proofs.}
Evaluation of natural language proofs is the most self-explanatory. Correct solutions are those that comprise the mathematical steps necessary to reach the conclusion of the statement at hand, with no arrival at statements without proof, except for steps that follow simply from standard methods within the field. This is, as all human verification of natural language proofs, susceptible to human error, as discussed in \Cref{section:the-trust-spectrum}.

\textbf{CAS.}
CAS code has been evaluated in the same way as a natural language proof. To be classified as correct, it should reproduce all of the steps necessary for a mathematical proof. Code that verified only part of the statements of the natural language proof, or had a small number of seemingly recoverable errors has been deemed partial, and other code, or code that does not compile has been labeled as having major issues.

\textbf{Lean.}
Lean was the only ITP used by the participants of the challenge. Its evaluation is much simpler than that of natural language proofs and CAS code. It requires only a semantic inspection of the definitions relating to the problem statement and object, and a check that the code compiles successfully, in addition that the code does not use statements without formal verification via the use of \texttt{sorry} or \texttt{axiom} statements.

Lean code that does not compile or where the definitions don't match the problem have been classified as having major issues. Code that does compile and matches a significant partial step towards a proof, or code that imports known theorems in the field without formal proof via the use of \texttt{sorry} or \texttt{axiom} statements have been classified as partial.

\section{Dataset}
\label{section:dataset}
\ash{expected: approx 1 page}

In this section, we explain the various quantitative aspects of the submissions received as part of the Ramanujan Challenge.

The Ramanujan Challenge received
228
submissions.

207 of the submissions clearly relate to a specific challenge problem, with 167 including solutions to the questions and the other 40 to the conjectures. Of these 207, 39 are older versions of later submissions to the same challenge problems by the same submitter. These are ignored in our analysis so that submissions with very little variation in their solution do not skew the data. We do not refer to these hereon.

\begin{wrapfigure}{L}{0.44\textwidth}
    \includegraphics[width=\linewidth]{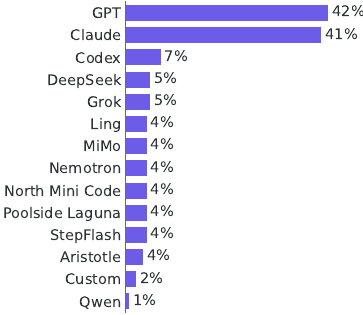}
    \vspace{-6pt}
    \caption{AI models used by the submissions, in percentages out of the total number of submissions, with data from the 61\% of submissions which reported their AI tools.}
    \label{figure:ai-used}
    \vspace{-8pt}
\end{wrapfigure}

\begin{figure}[ht]
    \centering
    \begin{minipage}[t]{0.48\textwidth}
        \centering
        \includegraphics[height=0.5\linewidth]{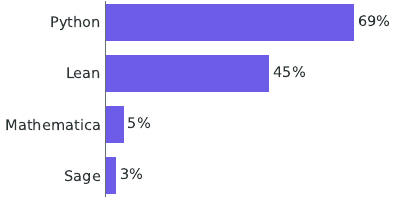}
        \caption{Computer Algebra Systems or Lean used by the submissions, in percentages out of the total number of submissions. Solutions to the cha}
        \label{figure:cas-used}
    \end{minipage}\hfill
    \begin{minipage}[t]{0.48\textwidth}
        \centering
        \includegraphics[height=0.5\textwidth,trim={0 48 0 16},clip]{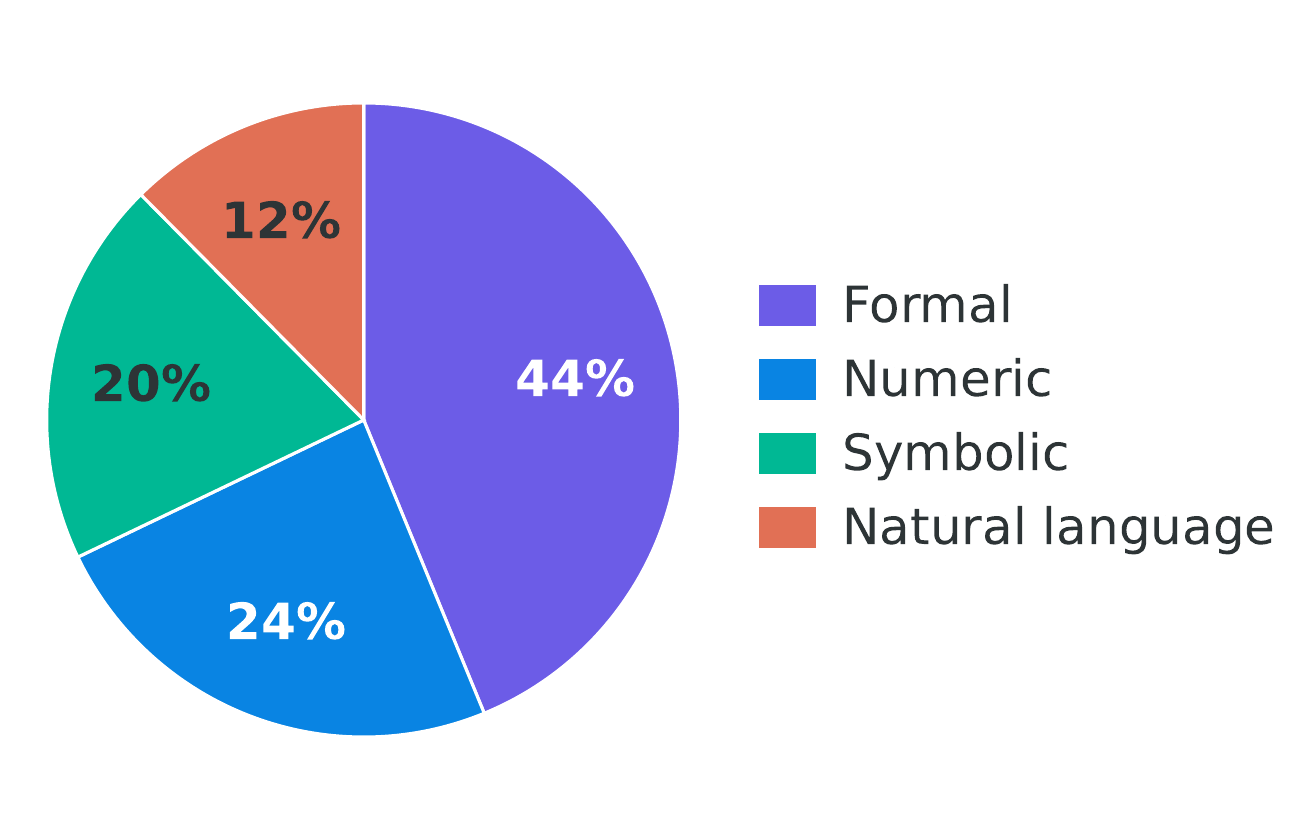}
        \caption{Type of formality of the submissions, in percentages out of the total number of submissions.}
        \label{figure:formality-levels}
    \end{minipage}
\end{figure}

Participants of the challenge used AI models, CAS and ITPs as listed in \Cref{figure:ai-used,figure:cas-used} respectively, with none using any ITPs other than Lean.

All submissions answering challenge questions supplied a natural language proof, with 88\% of them also including CAS or Lean code, as seen in \Cref{figure:cas-used}. 53 submissions to the challenge questions, and 4 to the conjectures, did not report the AI tools used in their submissions. We believe that submissions that did not specify the use of AI tools did not avoid their use, but rather only omitted the specification of said tools. We analyze these submissions as part of the data, with specific mention where it is excluded.

4 participants had only submissions that don't relate to a specific challenge problem. 9 participants submitted only 1 solution while 6 submitted solutions to all problems. The conjectures received 20 and 11 solutions respectively, with 10 people who submitted to both of them.

The type of formality of the solutions is detailed in \Cref{figure:formality-levels}. A solution was classified as having the lowest type of formality that it relied on, according to \Cref{section:the-trust-spectrum}, other than ``Human'' which was only given to solutions which included only a natural language proof, as the proofs are given a separate verdict from the code.

\WFskip

\section{Results}
\label{section:results}
\ash{Expected: approx 1.5 page}
\alan{min 1 page}

We now present the results of evaluating the submissions to the Ramanujan Challenge. In \Cref{subsection:results-challenge}, we detail our primary quantitative findings, including evaluation of the correctness of the solutions, similarity metrics and novelty analysis. \Cref{subsection:results-open-problems} details our observations regarding the open challenge questions. \Cref{subsection:pipeline-analysis} details the successes of our AI-assisted evaluation pipeline.

\begin{table}[ht]
    \caption{Human evaluation of the submissions, with each submission receiving a trinary grade for its natural language proof, CAS and Lean code, when present.}
    \label{tab:verdicts}
    \centering
    \small
    \begin{tabular}{cc:ccc:cccc}
        \multicolumn{2}{c}{} & \multicolumn{3}{c}{\textbf{Natural language proof}} & \multicolumn{4}{c}{\textbf{Computer Algebra System/Lean}} \\
        \\[-1.0ex]
        \multicolumn{1}{l}{Question} & \shortstack{Number of\\submissions} & Correct & Partial & \shortstack{Major\\issues} & Correct & Partial & \shortstack{Major\\issues} & \shortstack{No\\code} \\
        \hline
        1 & 23 & 11 & 6 & 6 & 7 & 11 & 3 & 2 \\
        2 & 16 & 7 & 3 & 6 & 2 & 8 & 4 & 2 \\
        3 & 19 & 10 & 3 & 6 & 5 & 7 & 5 & 2 \\
        4 & 19 & 4 & 7 & 8 & 1 & 13 & 3 & 2 \\
        5 & 11 & 2 & 2 & 7 & 0 & 5 & 4 & 2 \\
        6 & 20 & 10 & 5 & 5 & 6 & 9 & 2 & 3 \\
        7 & 15 & 6 & 2 & 7 & 4 & 6 & 3 & 2 \\
        8 & 14 & 4 & 2 & 8 & 0 & 6 & 6 & 2 \\
        \hline
        Total & 137 & 54 & 30 & 53 & 26 & 64 & 30 & 17 \\
    \end{tabular}
\end{table}

\subsection{Analysis of the Solutions to the Questions}
\label{subsection:results-challenge}

\begin{wrapfigure}[31]{R}{0.5\textwidth}
     \centering
     \includegraphics[width=0.85\linewidth]{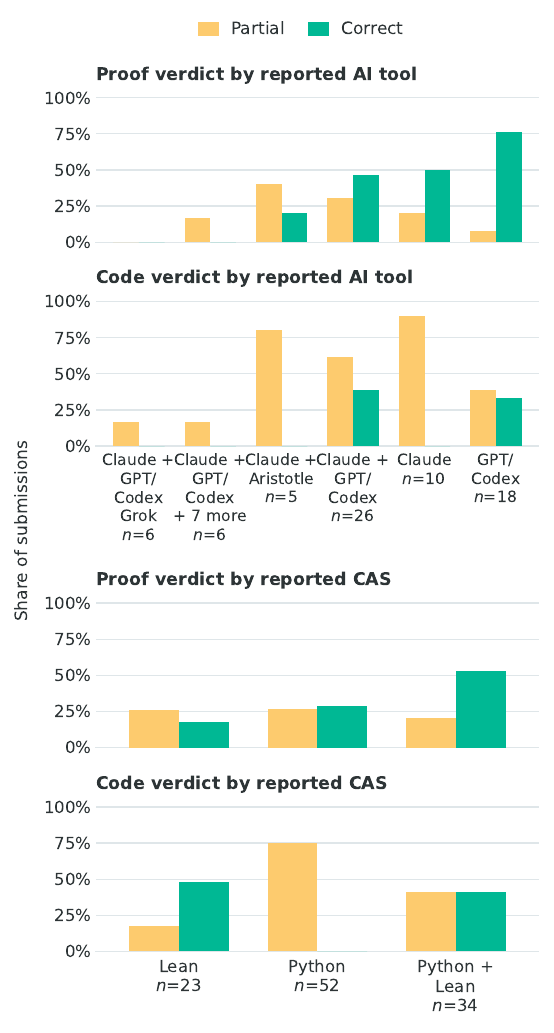}
     \vspace{-8pt}
     \caption{Correct and Partial verdicts for submissions to the challenge questions, by tool combination, for combinations that appeared in at least 5 submissions.}
     \label{figure:success-rates}
\end{wrapfigure}

\Cref{tab:verdicts,figure:success-rates} show the verdicts and tool-based success rates in submissions to the questions, as measured by the human reviewer separately for the natural language proof and for the verification code.

Of the reported AI tools, GPT/Codex achieved highest correctness for natural language proofs, while a combination of Claude and GPT or Codex achieved highest correctness for CAS/Lean code.

\begin{figure}[ht]
     \centering
     \includegraphics[width=0.9\linewidth]{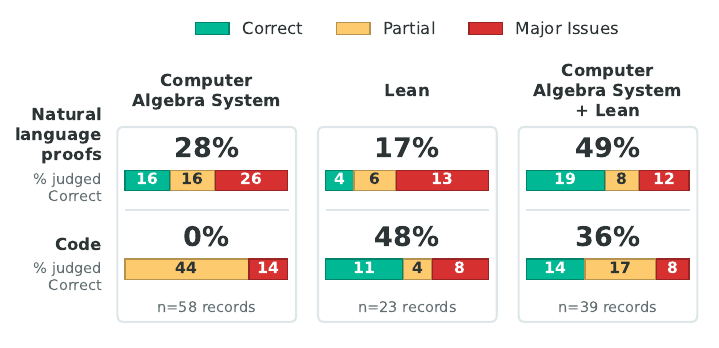}
     \caption{Correctness of natural language proofs and CAS derivation or Lean proofs, according to human evaluation. CAS code was all found to be at most partial and did not include comprehensive verification or proof of the problem statements, explaining the 0\% on the first column.}
     \label{figure:correctness-by-cas}
\end{figure}

Submissions including only natural language proofs achieved the highest correctness for natural language proofs; these were produced by 4 different submitters. This is followed by submissions using Python in combination with Lean, and highest code correctness was achieved by Lean alone. In general, natural language proofs in submissions that included both CAS and Lean code obtained the highest correctness across natural language proofs, and those including Lean code alone had the highest percentage of code correctness, both seen in \Cref{figure:correctness-by-cas}.

\begin{figure}[ht]
     \centering
     \includegraphics[width=0.9\linewidth]{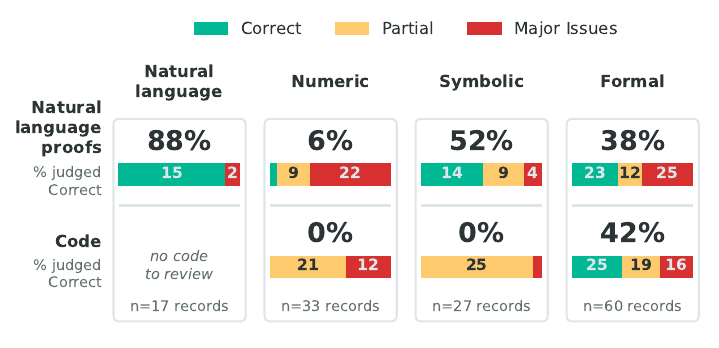}
     \caption{Correctness of natural language proofs and CAS/Lean code based on their level of formality. Code which did not include a Lean formalization did not make complete verification or proof and no such code was judged as correct.}
     \label{figure:correctness-by-trust}
\end{figure}

Viewing correctness by level of formality in \Cref{figure:correctness-by-trust} reinforces the usefulness of different levels of formality; Lean formalizations alone achieved correctness in code, but symbolic verification accompanying the most correct natural language proofs. Among solutions with code, those with symbolic verification had the highest natural language correctness of 52\%, above those including formal verification at 38\% and well above the 6\% of numerical checks. The highest rate belongs to the 17 solutions without code. We note that 15 of these come from only two submitters, reflecting more on those submitters than on the level of formality.

\begin{figure}[ht]
     \centering
     \includegraphics[width=0.7\linewidth]{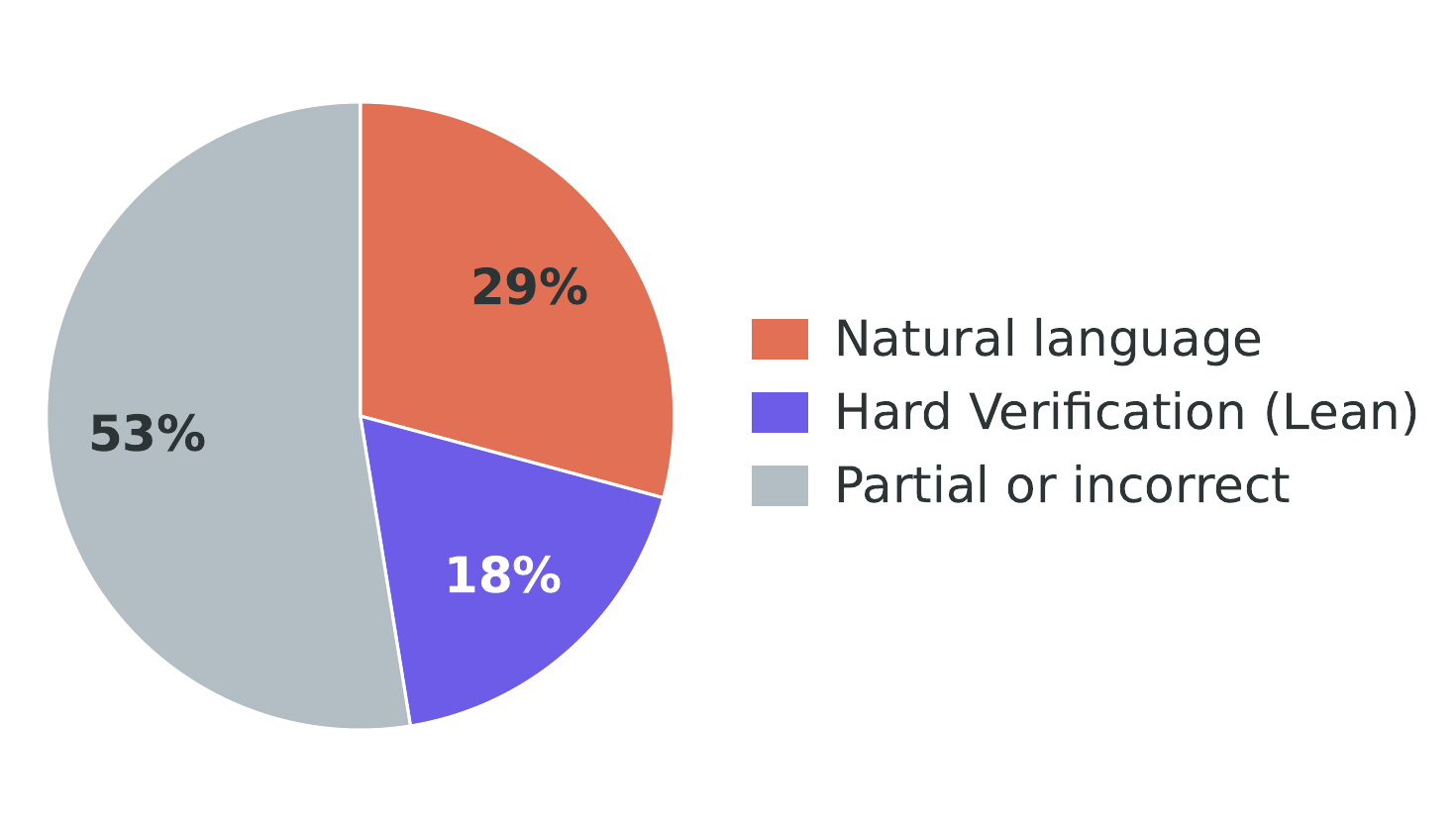}
     \caption{Trust level in the solutions to the challenge questions, in percentages.}
     \label{figure:trust-levels}
\end{figure}

\begin{figure}[ht]
     \centering
     \includegraphics[width=0.8\linewidth]{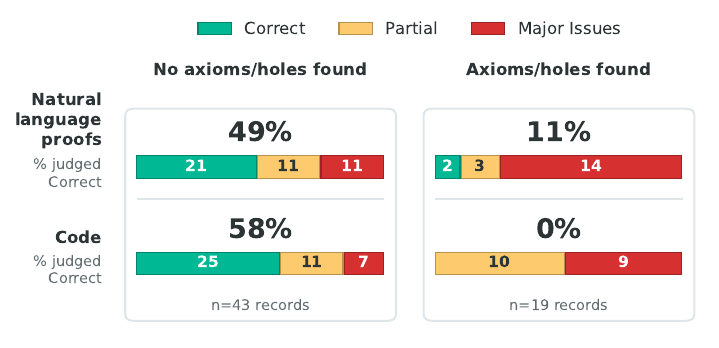}
     \caption{Correctness of natural-language proofs and code for the submissions that used Lean, split by whether their submitted Lean source contains \texttt{axiom}, \texttt{constant}, \texttt{sorry}, or \texttt{admit}.}
     \label{figure:correctness-by-lean-assumptions}
\end{figure}

Some Lean formalizations relied on certain statements without proof, via use of statements such as \texttt{axiom}, \texttt{constant}, \texttt{sorry}, or \texttt{admit}. Correctness of such formalizations, and the natural language proofs in their submission, are both significantly reduced by the use of these statements, as can be seen in \Cref{figure:correctness-by-lean-assumptions}. None of the code including formalizations with these statements was judged to be correct, compared to 58\% of code in the rest of submissions which included Lean code. We see also that only 11\% of natural language solutions accompanying such formalizations were found to be correct, with a higher 49\% for natural language proofs coming with Lean formalizations not including such statements. This demonstrates a correlation between correctness of the natural language proof and completeness of the Lean formalization.

Despite the advantage in the use of Lean for achieving a correct solution in either natural language or code, the first correct solutions to 3 of the challenge questions included a correct natural language proof, with no Lean formalization. The first correct solutions to challenge questions 3, 6 and 7 did not include Lean formalizations. Still, the first correct solutions to 5 of the questions did include correct Lean formalizations, in spite of the lower overall percentage of such solutions (which can be seen in \Cref{figure:trust-levels}).

This figure also shows that publicly available AI tools at the time of the challenge still struggle to consistently produce correct, semantically verifiable solutions and formalizations for the challenge problems.

\subsection{Solutions to the Conjectures}
\label{subsection:results-open-problems}

We present here the results regarding the two conjectures of the challenge, which are described in \Cref{appendix:challenge-conjectures}. \Cref{conjecture:knot-polynomial} received solutions including a correct Lean-formalized proof and multiple other proofs that were deemed likely correct after discussion with the problem author. Among the evaluated submissions, \Cref{conjecture:zeta3-optimality} received only partial solutions, and it remains open. We discuss these in more detail.

Due to scope and time constraints, the solutions to the conjectures have not yet been fully evaluated. Their evaluation will be included in the coming weeks.

The solutions to \Cref{conjecture:knot-polynomial} that were evaluated were those presenting a solution in Lean, and partially those which only gave a natural language proof.
Discussion with the author of the problem suggests that multiple solutions show good directions towards a proof. Evaluation of the Lean files for this problem shows one correct and complete formal proof, with two of the other three solutions having major issues and not formalizing the theorem statement, and one providing proof only for partial results.

Evaluation of the earlier solutions to \Cref{conjecture:zeta3-optimality} found only partial solutions or those with major issues, with all solutions admitting to only attempt proving partial results.



\subsection{Analysis of the Evaluation Pipeline}
\label{subsection:pipeline-analysis}

We analyze the evaluation pipeline as a method of increasing performance of LLM agents in the correct classification of challenge solutions. A Claude Opus 5 agent was used twice, once via one-shot prompting and once with the use of our evaluation pipeline, to classify the parts of the solution as correct, partial or with major issues. These were compared to human verdicts, which were considered the reference labels, with results detailed in \Cref{table:agreement-and-false-positives}.

Use of the pipeline increased overall agreement with the human verdict from 65.7\% to 66.4\% for natural language proofs and from 45.8\% to 58.3\% for code.

Without the pipeline, the agent was too strict on only 2.9\% of solutions when using one-shot prompting in the evaluation of natural language proofs, and on only 4.2\% of them for CAS or Lean code. It was too lax on 31.4\% of natural language proofs and 50.0\% of code, ignoring issues in them. 23.4\% of the natural language proofs, and 38.3\% of the code had been incorrectly classified as Correct.

These issues have been corrected with use of the pipeline, with reduction from 31.4\% to 18.2\% for natural language and 50.0\% to 12.5\% for code in the agent being too lax, and reduction from 23.4\% to 5.8\% for natural language and from 38.3\% to 3.3\% for code in false positives, less by factors of 4 and 11.5 respectively.

\ash{Sec 7 onwards = about 1.5 page}

\section{Limitations}
\label{section:limiations}
\ash{1 para: Club with limitations and scope presented in Section 3. Not an ICLR req to have this as a separate section}

There is uncertainty in the experiment results with regards to variation between participants and tools used. Challenge participants submitted solutions to as low as 1 and as high as all 10 challenge problems, potentially skewing the dataset. The participants were furthermore self-admitted to the challenge and chose and self-reported their use of tools for the solutions, introducing potential biases.

Evaluation of the submissions also leads to an amount of uncertainty, as they rely on human judgment, which is prone to error. Due to time and monetary constraints, evaluation of each natural language proof and CAS or Lean code was done by a single human evaluator, as described in \Cref{section:methodology} and without cross-checking. This leaves evaluation prone to human error and potentially skewed by the specific backgrounds of the evaluators.

\section{Discussion and Outlook}
\label{section:conclusion}
\ash{min of 2 paras}

As AI-assisted mathematics produces results at an increasing rate, reliable and scalable validation is becoming crucial.
Our analysis shows that a structured evaluation pipeline can improve judgment quality even when the underlying LLM is unchanged.
We expect this advantage to persist as models improve, particularly if evaluation pipelines evolve in parallel to model improvement.

The Ramanujan Challenge results show that research-level number theory problems are increasingly within reach of AI-assisted solvers.
Benchmarks at this level may therefore saturate as well, just as earlier benchmarks have,
motivating a continuous effort to construct up-to-date evaluations that track the advancing research frontier.
Open conjectures offer a renewable source of such problems, but because they lack known proofs, their use as benchmarks depends on reliable evaluation methods that do not require a priori ground truth to work well, such as the approach presented here.

Our analysis shows that natural language proofs supported by both Lean and CAS code were, on average, more successful than those relying on a single tool.
One plausible explanation is that the two forms of evidence are complementary: Lean guarantees the logical validity of the statement as formalized, whereas CAS can efficiently check intricate symbolic and numerical computations that are hard to formalize.
Together, they can therefore provide stronger evidence than either alone.
This suggests a broader role for CAS as a supporting tool for mathematical proofs. CAS certificates are often more readable than full formalizations while remaining machine-checkable, and tighter integration among CAS, proof assistants, and LLMs could turn verified computations into certified proof steps that combine readability with rigor.

More broadly, the setting considered here -- a natural language claim accompanied by code-based indirect evidence 
-- extends beyond number theory.
Similar situations arise throughout mathematics and increasingly in varied areas of AI for science, where generated claims may be supported by code, simulations, or data analyses.
As the volume of such results grows, systems capable of autonomously assessing the validity of a claim and the sufficiency of its supporting evidence may become increasingly crucial.  

\ash{Anybody else to be thanked?}

\section{AI Use}
\label{section:ai-use}
\ash{min of 2 paragraphs}

Our work relies heavily on the use of LLMs. The components where AI was majorly used include: 
\newline
\textbf{The dataset.}
All or most challenge solutions were generated by AI to some extent. A detailed analysis of this is provided in \Cref{section:dataset}. Since these are external contributions, the authors do not have access to the amount of tokens used for all submissions. \ash{provide more details in appendix if there is time.}
\newline
\textbf{Evaluation of solutions.} 
Codex with GPT-5.6-Sol with high reasoning effort and Claude Opus 5 with high reasoning effort were used heavily in the evaluation pipeline. Codex with GPT-5.6 was also used as an assistant during manual human evaluation of the solutions in addressing different questions regarding derivation gaps and proof outlines, with answers only inspected by the human reviewer as part of their own evaluation.
\newline
\textbf{Graphs.} 
Code was generated with the help of AI to analyze and create graphs.
\newline
\textbf{Survey.} 
LLMs were used for the purpose of literature survey.
\newline
\textbf{Grammar and phrasing.}
Feedback regarding correct grammar and phrasing.

\ash{add a table with tokens, model name and cost used for various tasks?}

The following was the contribution differentiated based on the models that were used:
\newline
\textbf{Claude.} We used a total of about 697,989,732 tokens over Opus 4.8, Opus 5, and Fable 5, with about 268,690,427 for the code summary files, about 132,790,138 for verdicts on the correctness of the solutions, and the rest for interactive sessions. We worked on a Claude Max subscription costing 100 USD per month for 3 months.
\newline
\textbf{Codex.} We used about 661,940,351 tokens, with 170,386,997 used for the natural language proof summary and 6,608,918 used in the summary comparison files. 484,944,436 tokens were used in the interactive sessions during human evaluation of the submissions.
\newline
We confirm not having used generative AI tools for any other tasks related to this manuscript, and take responsibility for the final content of this work, including text, claims or artifacts produced with the aid of generative AI.

\section{Ethics Statement}
\label{section:ethics}

\ash{taken from here: https://github.com/Henryhe09/ICLR2026-statement / maybe change a little so it isn't completely copied? :D}
 
We have taken care to avoid any biases or discriminatory outcomes in our research process. We have made sure to maintain the anonymity of the submitters to the Ramanujan Challenge.
We are committed to maintaining transparency and integrity throughout the research process.

\section{Reproducibility Statement}
\label{section:reproducibility}
Anonymized submission results are included in the supplemental materials. For each submission, the listed data includes the tools used in that submission, human evaluation results and claude evaluation results for that submission with and without the pipeline. The order of appearance in the file, within each problem of the challenge, is randomized so that the results and tools used cannot be inferred from the public list of submissions on the challenge website.

\bibliographystyle{iclr2027_conference}
\bibliography{library}

\newpage
\appendix

\section{The Challenge Questions}
\label[appendix]{appendix:challenge-questions}

\begin{question}[Polynomial continued fraction for $\pi$]
\label{question:pcf-pi}
Let:
\begin{equation}
\label{equation:coefficients}
a_n = -220n^3 - 484n^2 - 301n - 42,
\qquad
b_n = 4n^2(2n+1)^2(5n-4)(5n+6).
\end{equation}

Prove:
\begin{equation}
\label{eq:challenge_PCF}
a_0 + \cfrac{b_1}{a_1 + \cfrac{b_2}{a_2 + \cfrac{b_3}{a_3 + \ddots}}} \;=\; \frac{6}{3-\pi}.
\end{equation}
\end{question}

\begin{question}[Euler's constant $\gamma$ as an Ap\'ery limit]
\label{question:euler}
For $n\geq 0$, define:
\begin{align*}
0 = &\left(- 8 n^{3} - 51 n^{2} - 105 n - 68\right) u_n \\
& + \left(24 n^{5} + 337 n^{4} + 1833 n^{3} + 4818 n^{2} + 6092 n + 2928\right) u_{n-1} \\
&- \left(n + 2\right) \left(n + 3\right) \left(24 n^{5} + 273 n^{4} + 1150 n^{3} + 2154 n^{2} + 1635 n + 268\right) u_{n-2} \\
& +\left(n + 1\right) \left(n + 2\right)^{4} \left(n + 3\right) \left(8 n^{3} + 75 n^{2} + 231 n + 232\right) u_{n-3}\\
\end{align*}

Let $p_n, q_n$ be two solutions of the recurrence, defined by the initial values:
\[p_{-3} = 0, \quad p_{-2} = 7, \quad p_{-1} = 179\]
\[q_{-3} = 1, \quad q_{-2} = 12, \quad q_{-1} = 306\]

Prove
\[
\lim_{n\to\infty}{\frac{p_n}{q_n}}=\gamma,
\]
where $\gamma$ is Euler's constant.
\end{question}

\begin{question}[The sum $\pi+e$ as an Ap\'ery limit]
\label{question:pie}
For $n\geq 1,$ define
\begin{align*}
0 = & \left(- n^{3} + 2 n^{2} + 7 n + 3\right) u_n \\
&+ \left(n + 2\right) \left(2 n^{4} + n^{3} - 26 n^{2} - 48 n - 19\right) u_{n-1} \\
&+ \left(n + 2\right) \left(n^{6} + 9 n^{5} + 8 n^{4} - 87 n^{3} - 249 n^{2} - 234 n - 68\right) u_{n-2} \\
&+ \left(n + 1\right)^{2} \left(n + 2\right) \left(2 n^{5} + 3 n^{4} - 13 n^{3} - 21 n^{2} + 4\right) u_{n-3} \\
& - n^{3} \left(n + 1\right)^{2} \left(n + 2\right) \left(n^{3} + n^{2} - 8 n - 11\right) u_{n-4}.
\end{align*}

Let $p_n, q_n$ be two solutions of the recurrence, defined by the initial values:
\[p_{-3} = 1, \quad p_{-2} = 1, \quad p_{-1} = 20, \quad p_0 = 296\]
\[q_{-3} = 1, \quad q_{-2} = 0, \quad q_{-1} = 4, \quad q_0 = 48\]

Prove
\[
\lim_{n\to\infty}\frac{p_n}{q_n} = \pi + e.
\]
\end{question}

\begin{question}[Harmonic numbers, a polylogarithm and zeta values]
\label{question:harmonic}
Prove:
\begingroup
\small
\[
\sum_{m=0}^{\infty}\sum_{k=0}^{m} \frac{\binom{m}{k}^2 H_k^2}{(m+1)^2\binom{2m}{m}} =
\]
\[
20 \operatorname{Li}_4\!\left(\frac{1}{2}\right) + \frac{5}{6}\log^4(2) + 10 \zeta(2) - \frac{65}{9}\zeta(2)^2 -
 \log^2(2) (12 + 5 \zeta(2))
 + \frac{1}{2}\zeta(3)
 + \log(2)\left(\frac{35}{2}\zeta(3) - 16\right),
\]
\endgroup

where $H_0=0$ and for $k\geq 1$, $H_k$ is the $k$-th harmonic number.
\end{question}

\begin{question}[Efficient rational approximation of Catalan's constant $G$]
\label{question:catalan}
For $n\geq 0$, let
\[
M(n)=\bigl(m_{ij}(n)\bigr)_{1\leq i,j\leq 3},
\]
where
\[
\begin{aligned}
m_{11}(n)&=(-2n-5)(n+3)^2(136n^4+1424n^3+5548n^2+9551n+6141),\\
m_{12}(n)&=384n^6+6384n^5+44168n^4+162698n^3+336377n^2+369933n+169011,\\
m_{13}(n)&=-480n^4-4980n^3-19210n^2-32690n-20730,\\
m_{21}(n)&=(n+2)^2(n+3)^2(4n+10)(48n^3+386n^2+1017n+879),\\
m_{22}(n)&=(n+2)^2(-272n^5-3848n^4-21732n^3-61184n^2-85761n-47808),\\
m_{23}(n)&=(n+2)^2(320n^3+2540n^2+6610n+5640),\\
m_{31}(n)&=(-4n-10)(n+2)^2(n+3)^2(32n^4+302n^3+1037n^2+1530n+813),\\
m_{32}(n)&=(n+2)^2(192n^6+2984n^5+19116n^4+64452n^3+120256n^2+117279n+46476),\\
m_{33}(n)&=(n+2)^2(-16n^5-408n^4-2912n^3-8884n^2-12254n-6240).
\end{aligned}
\]

For $N\geq 0$, define
\[
\mathcal M_N
=
M(0)M(1)\cdots M(N-1).
\]

Choose initial conditions
\[
A=
\begin{pmatrix}
30921 & -32972 & 8240\\
33750 & -36000 & 9000
\end{pmatrix}.
\]
Write
\[
A\mathcal M_N
=
\begin{pmatrix}
P_{N,1} & P_{N,2} & P_{N,3}\\
Q_{N,1} & Q_{N,2} & Q_{N,3}
\end{pmatrix}.
\]
Then, for each $j=1,2,3$, prove:
\[
\lim_{N\to\infty}
\frac{P_{N,j}}{Q_{N,j}}
=
G,
\]
where
\[
G=\sum_{k=0}^{\infty}\frac{(-1)^k}{(2k+1)^2}
\]
is Catalan's constant.
\end{question}

\begin{question}[A series for $\zeta(2)+\zeta(3)$]
\label{question:zeta-series}
Let $(u_n)_{n\geq 1}$ be the sequence defined by
\[
u_1=-\frac{93}{4480},
\qquad
u_2=-\frac{117}{14000},
\]
and, for $n\geq 3$, by the recurrence
\begin{align*}
0=&-2(n+3)^3(2n+5)(3n+5)u_n \\
&\quad +(n+2)^2(15n^3+85n^2+155n+93)u_{n-1}\\
&\quad -(n+1)^3(n+2)(3n+8)u_{n-2}.
\end{align*}

Prove:
\[
\frac{2077}{720}+\sum_{j=1}^{\infty}u_j
=
\zeta(2)+\zeta(3).
\]
\end{question}

\begin{question}[Efficient four-term recurrence for $\zeta(2)+\zeta(3)$]
\label{question:zeta-fourterm}
For \(n\ge 0\), define

$A_n = 1024(2n+5)^4(2n+7)^3(2n+9)^3\bigl(946n^2+6407n+10860\bigr)$,\\
$B_n = 128(2n+7)^3(2n+9)^3\bigl(104060n^6+1745370n^5+12145238n^4+44886481n^3+92943995n^2+102256019n+46709052\bigr)$,\\
$C_n = 16(n+3)^4(2n+9)^3\bigl(3784n^5+57792n^4+351019n^3+1059230n^2+1587211n+944620\bigr)$,\\
$D_n = (n+3)^4(n+4)^6\bigl(946n^2+4515n+5399\bigr)$.

Let \((p_n)_{n\ge 0}\) and \((q_n)_{n\ge 0}\) be the two solutions of the recurrence
\[
u_{n+1} = \frac{B_n}{A_n}\,u_n-\frac{C_{n-1}}{A_{n-1}}\,u_{n-1}+\frac{D_{n-2}}{A_{n-2}}\,u_{n-2}, \qquad n\ge 2,
\]
with initial conditions
\[
p_0=-612218384750,\qquad
p_1=-\frac{9525021973931919}{18100},\qquad
p_2=-\frac{29561828382772029}{65380},
\]
and
\[
q_0=-215040420000,\qquad
q_1=-\frac{167282265043404}{905},\qquad
q_2=-\frac{964185327658080}{6071}.
\]

Prove:
\[
\lim_{n\to\infty}\frac{p_n}{q_n}=\zeta(2)+\zeta(3).
\]
\end{question}

\begin{question}[Very fast rational approximation of $\sqrt{10005}/\pi$]
\label{question:sqrt10005}
Let
\[
R=151931373056001,\qquad u=2n+3,\qquad
w=u(3u-2)(3u+2).
\]
For \(n\geq 0\), let
\[
M(n)=
\begin{pmatrix}
\dfrac{a_1}{w} & \dfrac{a_2}{w} & \dfrac{a_3}{w} & \dfrac{144R(u-1)^2}{w}\\[0.8em]
-u^3 & -3u^2 & -3u & -1\\[0.6em]
\dfrac{b_1}{144R} & -\dfrac{b_2}{72R} & -\dfrac{b_3}{36R} & -\dfrac{2u+2R-7}{2R}\\[0.8em]
\dfrac{c_1}{288R^2} & \dfrac{c_2}{144R^2} & \dfrac{c_3}{72R^2} & \dfrac{c_4}{4R^2}
\end{pmatrix},
\]
where
\[
\begin{aligned}
a_1&=(144R-99)u^5-(288R-333)u^4+(144R-229)u^3-114u^2+40u+64,\\
a_2&=(432R-243)u^4-(864R-909)u^3+(432R-868)u^2-80u+272,\\
a_3&=(432R-153)u^3-(864R-648)u^2+(432R-860)u+360,
\end{aligned}
\]
\[
\begin{aligned}
b_1&=9u^4-(144R-63)u^3+158u^2+168u+64,\\
b_2&=36u^3+(216R-189)u^2-316u-168,\\
b_3&=54u^2+(108R-189)u-158,
\end{aligned}
\]
and
\[
\begin{aligned}
c_1={}&18u^5+(54R+45)u^4-(288R^2-378R+251)u^3\\
&\quad +(948R-1086)u^2+(1008R-1384)u+(384R-576),\\
c_2={}&(153R-72)u^4-(657R-702)u^3-(432R^2-1292R+1069)u^2\\
&\quad +(2064R-2508)u+(1072R-1512),\\
c_3={}&(180R-108)u^3-(891R-864)u^2-(216R^2-1450R+1385)u\\
&\quad +(1116R-1422),\\
c_4={}&(6R-4)u^2-(33R-32)u-(4R^2-58R-236337691420383).
\end{aligned}
\]

For \(N\geq 0\), define
\[
\mathcal M_N=M(0)M(1)\cdots M(N-1),
\]
with the convention that \(\mathcal M_0=I\).

Choose initial conditions
\[
A=
\begin{pmatrix}
A_1&A_2&A_3&A_4\\
B_1&B_2&B_3&B_4
\end{pmatrix},
\]
where
\[
\begin{aligned}
(A_1,A_2,A_3,A_4)
={}&(
37169305760442252761441,\,
111507917281327441564208,\\
&\quad
111507917281327599720129,\,
37169305760442410917362),
\end{aligned}
\]
and
\[
\begin{aligned}
(B_1,B_2,B_3,B_4)
={}&(
1167416361542639692320,\,
3502249084627896132160,\\
&\quad
3502249084627879697280,\,
1167416361542622723840).
\end{aligned}
\]

Write
\[
A\mathcal M_N
=
\begin{pmatrix}
P_{N,1}&P_{N,2}&P_{N,3}&P_{N,4}\\
Q_{N,1}&Q_{N,2}&Q_{N,3}&Q_{N,4}
\end{pmatrix}.
\]
Then, for each \(j=1,2,3,4\):
\[
\lim_{N\to\infty}\frac{P_{N,j}}{Q_{N,j}}
=
\frac{\sqrt{10005}}{\pi}.
\]
\end{question}

\section{The Challenge Conjectures}
\label[appendix]{appendix:challenge-conjectures}
\ash{Merge with the previous section?}


This section contains open problems, presented here for the first time.
In each case, the statement is written in the form of an explicit formula so that it can be checked numerically. 

\begin{conjecture}[An integral over knot polynomial roots expressing \texorpdfstring{$\pi^2$}{π²}]
\label{conjecture:knot-polynomial}
Let $A_{7_{2}}(M, L)$ denote the $A$-polynomial for the prime knot $7_2$. 
\[
\begin{aligned}
A_{7_2}(M,L)
={}& L^5 + L^4\left(
M^{14}-M^{12}+3M^4+4M^2-2
\right) \\
&+ L^3\left(
-2M^{18}+5M^{16}+M^{14}-4M^{12}
+6M^8+5M^6+2M^4-4M^2+1
\right) \\
&+ L^2\left(
M^{22}-4M^{20}+2M^{18}+5M^{16}
+6M^{14}-4M^{10}+M^8+5M^6-2M^4
\right) \\
&+ L\left(
-2M^{22}+4M^{20}+3M^{18}-M^{10}+M^8
\right)+ M^{22}.
\end{aligned}
\]
Let $\alpha \approx 0.349269\ldots$ denote the real solution of $A_{7_{2}}\left( \alpha, \alpha^{1/2} \right) = 0$ closest to $0.349269$.
 Let $\beta \approx 0.406813\ldots$ denote the real solution of $A_{7_{2}}(\beta, \beta) = 0$ closest to $0.406813$
 Let $y = y(x)$ denote the curve satisfying $A(x, y(x)) = 0$ for $\alpha \leq x \leq \beta$ and such that $y(x)$ is positive and decreasing with $y'(x) \leq -2$. 
 Prove the following formula, which was discovered empirically: 
 $$ \frac{4 \pi ^2}{85} = \int_{\alpha}^{\beta} \left( \log x \frac{dy}{y} - \log y \frac{dx}{x} \right). $$

\end{conjecture}
This result is a surprisingly simple conjectural closed form of a Godbillon–Vey type Knot invariant \citep{Khoi2008}.

\begin{conjecture}[Optimality of Ap\'ery's irrationality-measure bound for \texorpdfstring{$\zeta(3)$}{zeta(3)}]
\label{conjecture:zeta3-optimality}

The sequences $(a_n)_{n\ge 0}$ and $(b_n)_{n\ge 0}$ are  solutions of Ap\'ery's recurrence
\[
  (n+1)^3 u_{n+1}
  -\bigl(34n^3+51n^2+27n+5\bigr)u_n
  +n^3 u_{n-1}=0
  \qquad (n\ge 1),
\]
with initial values
\[
  a_0=0,\quad a_1=6,
  \qquad
  b_0=1,\quad b_1=5.
\]
Define $d_n\coloneqq \operatorname{lcm}(1,\dots,n)^3$. Note that $d_na_n,d_nb_n\in \mathbb{Z}$ as shown in \citep{Apery1979}.

Prove:
\[
\operatorname{gcd}(d_na_n,d_nb_n) = e^{o(n)}.
\]
\end{conjecture}
The $a_n,b_n$ sequences were generated by Ap\'ery 
in his proof of the irrationality of $\zeta(3)$ \citep{Apery1979}.
Specifically, considering the linear forms $a_n-\zeta(3) b_n$ 
gives a bound on the irrationality measure of $\zeta(3)$, by
demonstrating that $d_n a_n,d_nb_n\in\mathbb{Z}$. 

The above conjecture shows that the irrationality-measure bound
found by Ap\'ery is tight for his particular formula.

\end{document}

%% file: math_commands.tex
\usepackage{amsmath,amsfonts,bm}

\def\eqref#1{equation~\ref{#1}}

\def\1{\bm{1}}

\DeclareMathAlphabet{\mathsfit}{\encodingdefault}{\sfdefault}{m}{sl}
\SetMathAlphabet{\mathsfit}{bold}{\encodingdefault}{\sfdefault}{bx}{n}



%% file: figures/manual_evaluation_color_diagram.tex
\definecolor{ink}{HTML}{2D3436}%
\definecolor{blue}{HTML}{0984E3}%
\definecolor{purple}{HTML}{6C5CE7}%
\definecolor{teal}{HTML}{E17055}%
\definecolor{amber}{HTML}{D63031}%
\definecolor{claudeaccent}{HTML}{E84393}%
\definecolor{gold}{HTML}{00B894}%
\definecolor{softgold}{HTML}{55EFC4}%
\definecolor{paper}{HTML}{FFEAA7}%
\definecolor{softink}{HTML}{FFEAA7}%
\definecolor{softblue}{HTML}{74B9FF}%
\definecolor{softpurple}{HTML}{A29BFE}%
\definecolor{softteal}{HTML}{FAB1A0}%
\definecolor{softamber}{HTML}{FF7675}%
\begin{tikzpicture}[x=1mm,y=1mm,scale=.97,transform shape,
  font=\sffamily\fontsize{7.5}{8.5}\selectfont,text=ink,
  card/.style={draw=#1!65!ink,fill=#1!9!white,rounded corners=1.5pt,
    line width=.5pt,align=center,inner sep=1mm},
  flow/.style={draw=ink,line width=.65pt,-{Stealth[length=1.5mm]}},
  head/.style={font=\sffamily\bfseries\fontsize{7.5}{9}\selectfont,align=center},
  pics/person/.style={code={
    \fill[teal!65!ink] (0,1) circle (1.1);
    \fill[teal!65!ink,rounded corners=.5pt] (-1.8,-2) rectangle (1.8,-.5);
  }},
  pics/bot/.style args={#1}{code={
    \draw[draw=#1!65!ink,fill=#1!20!white,rounded corners=.6pt,line width=.5pt]
      (-2,-1.5) rectangle (2,1.5);
    \fill[#1!65!ink] (-.8,.2) circle (.25);
    \fill[#1!65!ink] (.8,.2) circle (.25);
    \draw[#1!65!ink,line width=.5pt] (-.7,-.7)--(.7,-.7) (0,1.5)--(0,2.3);
    \fill[#1!65!ink] (0,2.4) circle (.3);
  }},
  pics/movingdoc/.style args={#1}{code={
    \draw[#1!45!white,line width=.5pt,line cap=round]
      (-2.4,1)--(-4.2,1) (-2.4,0)--(-5.2,0) (-2.4,-1)--(-4.2,-1);
    \pic {doc=#1};
  }},
  pics/doc/.style args={#1}{code={
    \draw[#1!65!ink,fill=#1!10!white,line width=.5pt] (-1.7,-2.2)--(1.7,-2.2)
      --(1.7,1.1)--(.6,2.2)--(-1.7,2.2)--cycle;
    \draw[#1!65!ink,line width=.4pt] (.6,2.2)--(.6,1.1)--(1.7,1.1)
      (-1,-.1)--(1,-.1) (-1,-.9)--(1,-.9);
  }}
]
\foreach \left/\right/\tone/\number/\title/\foreground in {
  0/18/gold/{}/{Submission}/ink,
  77/104/amber/2/{Summary\\Comparison}/ink,
  106/138/teal/3/{Human\\Evaluation}/ink}{
  \ifnum\pipelinevariant=1
    \filldraw[fill=\tone!3!white,draw=\tone!35!white,rounded corners=2pt,line width=.5pt]
      (\left,0) rectangle (\right,78);
    \fill[\tone!23!white,rounded corners=2pt] (\left,68) rectangle (\right,78);
    \node[head,text=\tone!45!ink] at ({(\left+\right)/2},73) {\number\,\title};
  \else
    \fill[\tone!55!ink,rounded corners=2pt] (\left,68) rectangle (\right,78);
    \node[head,text=white] at ({(\left+\right)/2},73) {\number\,\title};
    \draw[\tone!45!white,dashed,line width=.5pt] (\left,66)--(\left,0);
  \fi
}
\ifnum\pipelinevariant=1
  \filldraw[fill=purple!3!white,draw=purple!35!white,rounded corners=2pt,line width=.5pt]
    (20,40) rectangle (75,78);
  \fill[purple!23!white,rounded corners=2pt] (20,68) rectangle (75,78);
  \node[head,text=purple!45!ink] at (47.5,73)
    {1.1 Natural-Language Proof Summary};
  \filldraw[fill=claudeaccent!3!white,draw=claudeaccent!35!white,rounded corners=2pt,line width=.5pt]
    (20,0) rectangle (75,38);
  \fill[claudeaccent!23!white,rounded corners=2pt] (20,28) rectangle (75,38);
  \node[head,text=claudeaccent!45!ink] at (47.5,33) {1.2 Code Summary};
\else
  \fill[purple!55!ink,rounded corners=2pt] (20,68) rectangle (75,78);
  \node[head,text=white] at (47.5,73)
    {1.1 Natural-Language Proof Summary};
  \draw[purple!45!white,dashed,line width=.5pt] (20,66)--(20,40);
  \fill[claudeaccent!55!ink,rounded corners=2pt] (20,28) rectangle (75,38);
  \node[head,text=white] at (47.5,33) {1.2 Code Summary};
  \draw[claudeaccent!45!white,dashed,line width=.5pt] (20,26)--(20,0);
\fi
\node[card=gold,text width=13mm,minimum height=8mm] (pdf) at (9,52)
  {Natural Language\\Proof};
\node[card=gold,text width=13mm,minimum height=8mm] (cas) at (9,20)
  {Computer\\Algebra\\System};
\node[card=gold,text width=13mm,minimum height=8mm] (lean) at (9,8) {Lean Files};
\pic at (9,62) {doc=gold};
\node[card=blue,text width=20mm,minimum height=14mm] (basic) at (33.5,14)
  {\textbf{BasicGrade}\\[1mm]Python Metrics\\and Executions\\Results};
\draw[draw=ink,line width=.65pt] (cas.east)--(19,20)--(19,3.5)--(45,3.5)
  .. controls (46,3.5) and (45.5,14) .. (46.5,14);
\draw[flow] (19,17) .. controls (19,15.5) and (20.5,14) .. (basic.west);

\node[card=purple,text width=30mm,minimum height=12mm] (codex) at (47.5,52)
  {\textbf{Codex}\\Natural-Language Proof\\Summary and List of Issues};
\node[card=claudeaccent,text width=22mm,minimum height=15mm,anchor=north]
  (claude) at (60.5,21.5)
  {\textbf{Claude}\\Computer Algebra\\System and Lean\\Summary and\\List of Issues};
\pic at (35,62) {bot=purple};
\pic at (52,24.5) {bot=claudeaccent};
\pic at (71,62) {movingdoc=purple};
\pic at (71,24.7) {movingdoc=claudeaccent};
\draw[flow] (pdf.east)--(codex.west);
\draw[draw=ink,line width=.65pt] (lean.east)
  .. controls (18,8) and (19,7.5) .. (19,6);
\draw[draw=ink,line width=.65pt] (basic.east)--(46.5,14);
\draw[flow] (46.5,14)--(claude.west|-0,14);

\node[card=amber,text width=20mm,minimum height=16mm] (compare) at (90.5,34)
  {\textbf{Codex}\\Compare the\\Two Summaries};
\pic at (83,45.5) {bot=amber};
\pic at (87.5,21.5) {doc=purple};
\pic at (93.5,21.5) {doc=claudeaccent};
\pic at (100,45.5) {movingdoc=amber};
\draw[flow] (73,52) .. controls (76,52) and (76,51) .. (76,49)--(76,37)
  .. controls (76,35.5) and (77.5,34) .. (compare.west);
\draw[flow] (claude.east|-0,14) .. controls (76,14) and (76,15) .. (76,17)--(76,31)
  .. controls (76,32.5) and (77.5,34) .. (compare.west);

\node[card=teal,text width=25mm,minimum height=13mm] (human) at (122,49)
  {\textbf{Manual Evaluation}\\With Codex Assistance};
\pic at (122,62) {person};
\pic at (110,62) {doc=gold};
\pic at (116,62) {doc=purple};
\pic at (128,62) {doc=claudeaccent};
\pic at (134,62) {doc=amber};
\draw[flow,rounded corners=2mm] (9,64.4)--(9,66.5)--(105,66.5)--(105,46)--(human.west|-0,46);
\draw[flow,rounded corners=2mm] (claude.east|-0,14)--(105,14)--(105,46)--(human.west|-0,46);
\draw[draw=ink,line width=.65pt] (codex.east)--(98,52)
  .. controls (103,52) and (105,51.5) .. (105,49);
\draw[draw=ink,line width=.65pt] (compare.east)
  .. controls (104,34) and (105,35) .. (105,38);
\node[card=teal,text width=22mm,minimum height=8mm,
  font=\sffamily\fontsize{7}{8}\selectfont] (report) at (123,27)
  {Natural Language\\Proof Grade};
\node[card=teal,text width=22mm,minimum height=8mm,
  font=\sffamily\fontsize{7}{8}\selectfont] (rating) at (123,10)
  {Computer Algebra\\System/Lean Grade};
\draw[draw=ink,line width=.65pt] (human.south)--(122,38)--(108,38)--(108,10);
\draw[flow] (108,27)--(report.west);
\draw[flow] (108,10)--(rating.west);
\end{tikzpicture}